# American Sign Language Alphabet Recognition using Deep Learning

Nikhil Kasukurthi[1]     Brij Rokad[2]     Shiv Bidani[3]     Aju Dennisan[4]

{ [1]nikhil.kasukurthi, [2]brij.rokad, [3]shivbidani }@gmail.com    |    [4]daju@vit.ac.in

VIT University

*Abstract*- **Tremendous headway has been made in the field of 3D hand pose estimation but the 3D depth cameras are usually inaccessible. We propose a model to recognize American Sign Language alphabet from RGB images. Images for the training were resized and pre-processed before training the Deep Neural Network. The model was trained on a squeezenet architecture to make it capable of running on mobile devices with an accuracy of 83.29%.**

## 1. Introduction

Sign language conversion has been a long standing computer vision problem[1]. Several solutions have come up but none of them have been portable for them to be used in a standalone device or application. We plan on alleviating this problem by harnessing the power of the mobile phone and the recent advances in deep learning.

With the advent of deep learning, end to end models are being built for a wide range of problems that only require the images as input. Datasets have made it possible to harness the power of the models better. Imagenet is the best example, it is still driving innovation and advancements in computer vision. Another such dataset is the Microsoft Coco which is one of the benchmarks for image segmentation and human pose estimation. The problem of image classification has become very trivial now, on the other hand image segmentation is still quite difficult.

We propose an end to end solution that would require only a 2D image as an input. For this we follow a 3 Segment approach inspired by [2]. Our aim is to make it easy for people to communicate using the model. There are numerous people who use the ASL around the world. A vision based approach to our solution attempts to reduce the requirement of human translators and increase dependency on the user's phone for translation.

## 2. Related Work

There have been multiple instances where the accuracy of the Hidden Markov Model approaches 94% [3]. The HMM model utilizes 2 cameras. One camera is mounted on the desk (92%) and the other is a wearable cap with an attached camera (94%). The accuracy increases when the position of camera is changed. This shows that perspective plays an important role in the determination of the accuracy.

In [4] the system utilizes a PCA analyzer model to determine the position of the hand shape. It also uses a motion chain code to understand the hand movement to make determinations about signs that require movement. They have managed to achieve an error rate of 10.91%.

[5] employs two models, one an effective hierarchical feature characterization scheme and the other is a TMDHMM network to fasten the recognition process without any loss in recognition accuracy. In [6] there is a three-stage model that relies on movement of the hand. In the first phase the image of the hand is captured in a sequence. The sequence is evaluated to determine the events. These events determine the complete start to end positions. The second phase is the segmentation of the image and includes the training based on the contours of the hand. The final phase is the determination phase where the image is provided with a score to determine if it belongs to the signs that have been learned.

While [7] employs a wearable glove that is used on the hand with colors that assist the camera to determine the tips of the fingers as they are far apart on the color spectrum. The process involves reading the frame, segmenting the color, finding the centroid of the image and the final step is the determination of the result.

[8] has utilized a number of parameters including posture of the hand, position, motion and orientation. The model uses a typical HMM with a sub-par accuracy level due to the processing happening for 51 postures, 6 orientations and 6 positions. The accuracy obtained is around 80.4%.

Neural nets and Hough Transform have been used for ASL detection in certain models [9]. It utilizes a vector feature function for comparison. The vector feature is not prone to any disturbances due to rotation and scaling, this enables the system to be highly flexible and robust. The accuracy attained by this methodology was around 92% which shows that it is an industry standard solution to the ASL detection problem.

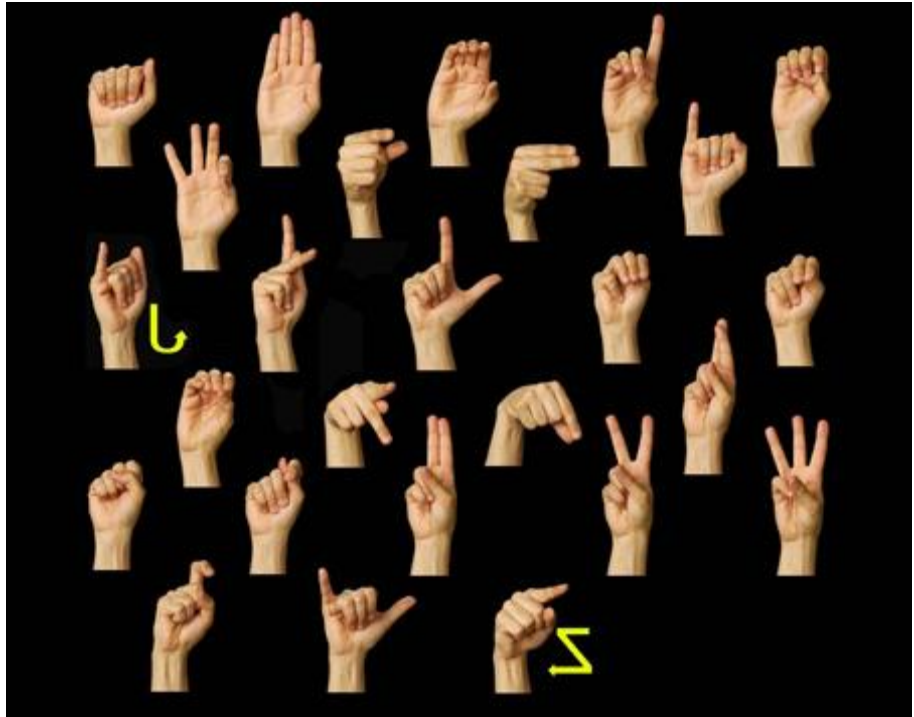

Fig. 1 ASL alphabet. Top Left - A

There are multiple other researchers that have aimed to solve the ASL detection problem with accuracies ranging from 70-94%.

While in [11] communication via gestures (ISL) comprises of static and in addition dynamic hand motions. A large portion of the ISL signals are created utilizing the two hands. A video database is made and used which contains a few recordings, for an expansive number of signs. Bearing histogram is the component utilized for arrangement because of its allure for brightening and introduction invariance. Two distinctive methodologies used for acknowledgment are Euclidean separation and K-closest neighbor measurements.

[12] Exhibits the assessment of different pixel level highlights for the dual handed sign language dataset. The element extraction techniques are Histogram of Orientation Gradient (HOG), Histogram of Boundary Description (HBD) and the Histogram of Edge Frequency (HOEF). The exactness of HOG and HBD found up to 71.4% and 77.3% while the precision of HOEF, all things considered, informational collection is 97.3% and in perfect condition 98.1%.

In [13] a wavelet based video division procedure is proposed which identifies states of different hand signs and head development in video based setup. Shape highlights of hand motions are extricated using elliptical Fourier descriptions which to the most elevated degree lessens the element vectors for a picture. The exploratory outcomes demonstrate that framework has an accuracy rate of 96%.

[14] Horn Schunck optical stream calculation removes highlights of the two hands giving position vectors of hands in each edge. The joined component framework having shape highlights prepare the back-propagation neural network. The ordered signs are mapped to content from the objective grid of ANN and changing over those content contributions to voice charges with windows content to application programmable interface. The word coordinating score over various occurrences of preparing and testing of the neural network brought around 90%.

Table 1. Comparison Table for Various SL Detection Models

| Author | Model / Method | Input | Feature Vector | Classification | Accuracy |
|---|---|---|---|---|---|
| Nandy, A., Prasad, et al[11] | HMM | Real time video | Orientation Histogram | Euclidean distance | 90% |
| Lilha, H., & Shivmurthy, D[12] | Histogram of Orientation Gradient (HOG), Histogram of Boundary Description (HBD) and the Histogram of Edge Frequency (HOEF) | Images | Histogram of Edge Frequency (HOEF). | Support Vector Machine | 98.1% |
| Kishore, P. V. V., & Kumar, P. R[13] | PCA | Video | Elliptical Fourier descriptors | Fuzzy | 91% |
| Kishore, P. V. V., Prasad, M. V. D., et al[14] | HMM | Real-time | Texture features | ANN- error back propagation algorithm | 91% |

## 3. Methodology

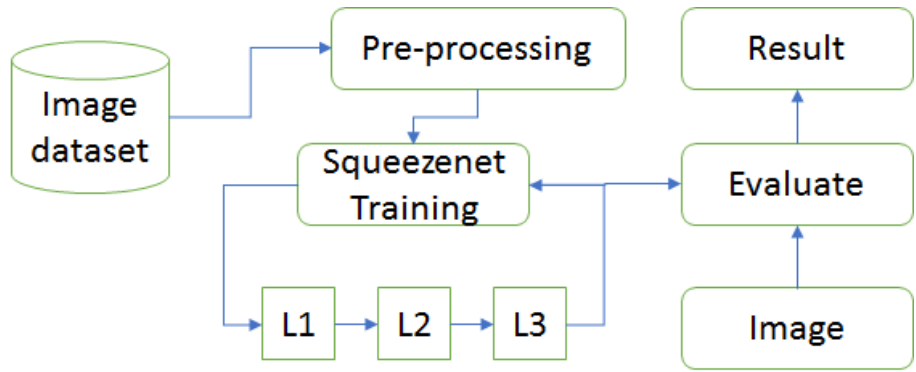

Fig. 2 Proposed Architecture

For translating the image to the relevant alphabet, we have trained the pre-trained model on the SqueezeNet model. Our model is trained on the Surrey Finger [10]. The trained model is then used for inference from the images that are being fed as input to the image. In accordance to the pre-trained model we process the image by dividing every pixel by 255 and then resizing the image to 244X244.

We have removed the top layers of the original network and add two dense layers followed by a softmax function that generated the output probabilities. A Stochastic Gradient Descent optimizer is used to ensure that the weight updates are not too abrupt and remain similar to the original pre-trained model. The loss function that we are using is the categorical cross entropy which takes in a loss over the target label and entire set of labels.

The cross-entropy function is used to determine the loss incurred by the network and is given by:

$$L(\theta) = -\frac{1}{n}\sum_i^n [y_i \log(p_i) + (1-y_i)\log(1-p_i)] = -\frac{1}{n}\sum_i^n \sum_j^m y_{ij}\log(p_{ij})$$

### 3.1 Dataset

The surrey finger dataset [10] has been used to train our model. It contains 41258 training and 2728 testing samples. Each sample provides RGB image (320x320 pixels), Depth map (320x320 pixels), Segmentation masks (320x320 pixels) for the classes: background, person, three classes for each finger and one for each palm, 21 Key points for each hand with their uv coordinates in the image frame, xyz coordinates in the world frame and a visibility indicator and Intrinsic Camera Matrix K. For our scenario, we use only the RGB images.

### 3.2 Image Preprocessing

To match the original training of the pretrained model of squeezenet, the mean value pixels are subtracted from all the images. Then resize the image to 244X244, to create more training data augmentation was applied. The data was shuffled in order to have a diverse sub dataset when picked randomly.

### 3.3 Squeezenet

The architecture that aim to follow is the Squeezenet architecture[15]. The squeezenet architecture comprises of a number of filters. The first level comprises of four 1x1 filters that are concatenated at the next layer. The concatenation ensures that the number of parameters is minimal. The primary objective of the squeezenet architecture is to reduce the number of parameters, and in turn the size of the network.

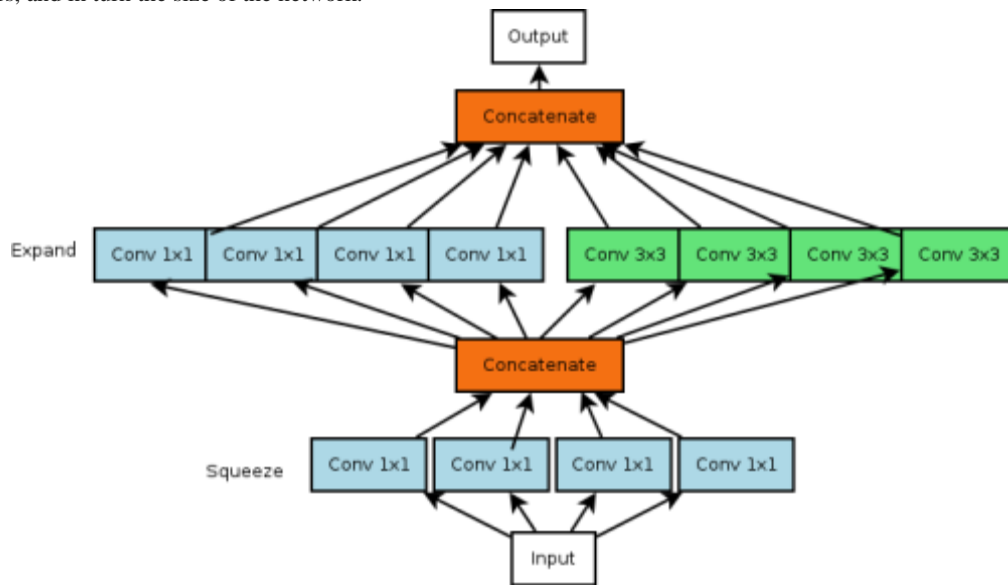

Fig. 3 Squeezenet Architecture

The concatenated layer is fed onto the expand layer and hence the number of interconnections between the squeeze layer and the expand layer are minimal. This ensures that the size of the network is low. The expand layer comprises of 3x3 filters along with more 1x1 filters. These are concatenated in order to attain the result.

### 3.4 Testing

The weights of the trained model are updated through training it over multiple epochs. The weights and biases of the network are trained over the epochs to determine a final network. This net is used to evaluate the input image. The evaluation is a low computation process and can be carried out on a handheld mobile device.

## 4. Experimental Results

Table 2. Test Images and Prediction

| Test Images | 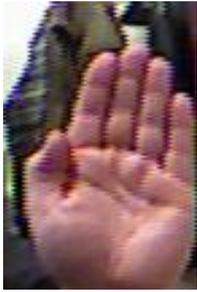 | 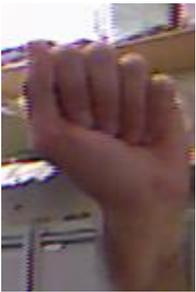 | 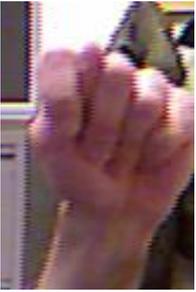 | 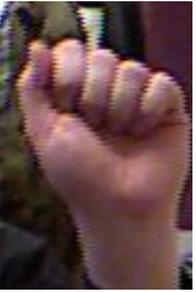 |
|---|---|---|---|---|
| Label | b | a | t | a |
| Prediction | b | a | a | a |

The model was trained on NVIDIA K80 GPU for a dataset size of 41,258 images. The model was trained for 10 epochs. The initial training accuracy and validation accuracy increases drastically till the 5th epoch. The accuracy then attains a plateau limit as the epochs increase. The maximum validation accuracy attained is 83.29% at the 9th epoch. Whereas the maximum training accuracy attained is 87.47%. The correlation between the training and validation accuracy is 98.47% which signifies that the model has been trained accurately.

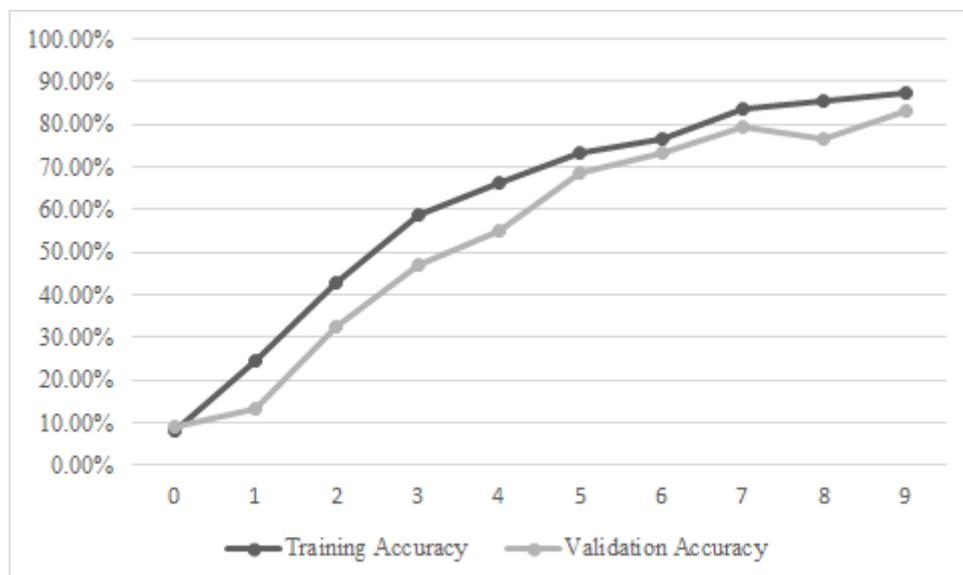

Fig. 5 Model Training Accuracy

In table 2 we show some of our results. The model is able to give accurate predictions but there are certain cases when it fails. From our observation, we noticed that this happens when similar looking alphabet like 'a' and 't' where the difference between them is thumb on the side for 'a' whereas 't' has thumb in between index and middle finger. When an image with different light conditions is given, or the fingers are not visible then it leads to a false prediction.

## 5. Conclusion

From the tests a maximum training accuracy of 87.47% is attained. The validation accuracy attained is 83.29%. The network learned the ASL that allowed it to predict the sign language in real time. The model developed was a squeezenet architecture which enabled the complete architecture to be stored on a mobile device. This helped with the accessibility of such a solution for the public. Hence, algorithmic recognition for mobile devices is currently preferred in order to present the majority of people with a highly accessible solution. In the future, the dataset preprocessing will help to improve the accuracy of the model. The lighting conditions and distance of the image from the camera should not affect the outcome of the prediction.